\documentclass[letterpaper, 10 pt, conference]{ieeeconf}  

\IEEEoverridecommandlockouts                              

\usepackage{graphics} 
\usepackage{epsfig} 
\usepackage{mathptmx} 
\usepackage{times} 
\usepackage{amsmath} 
\usepackage{amssymb}  
\usepackage{caption}
\usepackage{subcaption}
\usepackage{array}
\usepackage{siunitx}
\usepackage{tabularx}
\usepackage[T1]{fontenc}
\usepackage{makecell}
\usepackage{booktabs}
\usepackage{tabularx}
\usepackage[colorinlistoftodos]{todonotes}

\title{\LARGE \bf
From CAD to POMDP: Probabilistic Planning for Robotic Disassembly of End-of-Life Products
}

\author{Jan Baumgärtner$^{1*}$, Malte Hansjosten$^{1*}$,  David Hald$^{1*}$,\\ Adrian Hauptmannl$^{1}$, Alexander Puchta$^{1}$ and Jürgen Fleischer$^{1}$
\thanks{*These authors contributed equally to this work.}
\thanks{$^{1}$The authors are with the wbk Institute of Production Science,
 Karlsruhe Institute of Technology, 76131 Karlsruhe, Germany
 {\tt\small jan.baumgaertner@kit.edu}}%
\thanks{Funded by the Deutsche Forschungsgemeinschaft (DFG, German
Research Foundation) - SFB-1574 - 471687386. Thank you to Till Krause for providing the alternative models for the electric motor.
}
}

\begin{document}

\maketitle
\thispagestyle{empty}
\pagestyle{empty}

\begin{abstract}
To support the circular economy, robotic systems must not only assemble new products but also disassemble end-of-life (EOL) ones for reuse, recycling, or safe disposal. Existing approaches to disassembly sequence planning often assume deterministic and fully observable product models, yet real EOL products frequently deviate from their initial designs due to wear, corrosion, or undocumented repairs. We argue that disassembly should therefore be formulated as a Partially Observable Markov Decision Process (POMDP), which naturally captures uncertainty about the product's internal state.  
We present a mathematical formulation of disassembly as a POMDP, in which hidden variables represent uncertain structural or physical properties. Building on this formulation, we propose a task and motion planning framework that automatically derives specific POMDP models from CAD data, robot capabilities, and inspection results.  
To obtain tractable policies, we approximate this formulation with a reinforcement-learning approach that operates on stochastic action outcomes informed by inspection priors, while a Bayesian filter continuously maintains beliefs over latent EOL conditions during execution.  
Using three products on two robotic systems, we demonstrate that this probabilistic planning framework outperforms deterministic baselines in terms of average disassembly time and variance, generalizes across different robot setups, and successfully adapts to deviations from the CAD model, such as missing or stuck parts.
\end{abstract}

\section{INTRODUCTION}
\begin{figure}[h]
\centering
\includegraphics[width=\columnwidth]{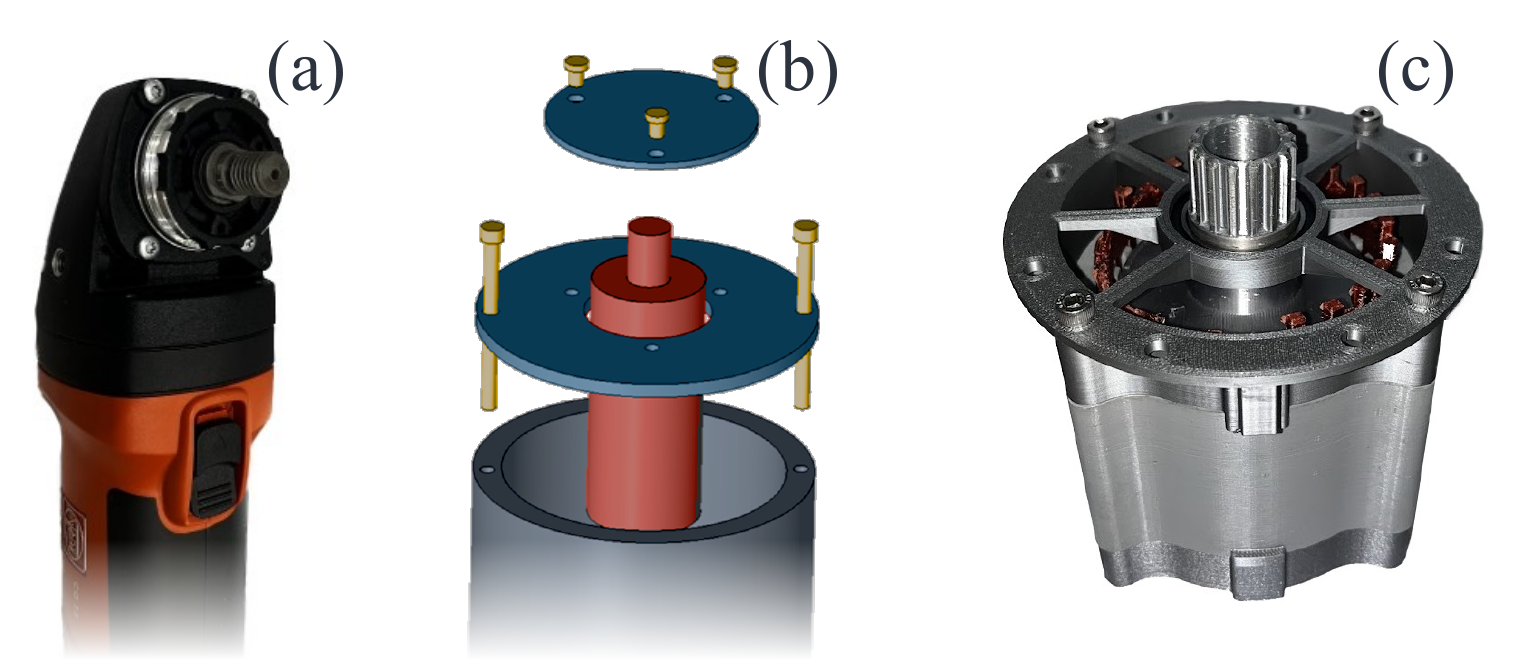}
\caption{Products used in the validation of the system.  (a) angle grinder, (b) electric motor with two lids, and (c) electric motor with a single lid.}
\label{fig:bg3-motor}
\end{figure}

Modern industrial production still follows a linear model of make-use-dispose, accelerating the depletion of natural resources on our planet. Circular economy approaches instead rely on recovering value from end-of-life (EOL) products through reuse, remanufacturing, or recycling. All of these processes require at least partial disassembly of products into their components, which is currently done mostly by hand.
Robotic disassembly can meet this challenge by offering scalable, safe, and economically viable alternatives to manual work.
The key challenge here is that end-of-life products often deviate significantly from their original CAD models due to wear, corrosion, damage, or undocumented repairs.
Consider the disassembly of the electric motor shown in Fig.~\ref{fig:bg3-motor}(b). If we want to extract the rotor (red), we can either remove the larger lid or the smaller lid. A deterministic planner will always prefer the larger lid, as it requires fewer screws to be removed. However, in practice, we might know that prior experience or optical inspection suggests that the larger screws are likely rusted shut.
A robust planner should be able to reason about such uncertainties to choose actions that are most likely to lead to a fast and successful disassembly instead of blindly assuming the best-case scenario. To solve this issue, we present a Partially Observable Markov Decision Process (POMDP)~\cite{cassandra1998survey}  formulation of robotic disassembly for EOL products, where hidden variables represent uncertain structural or physical properties (e.g., whether a screw is rusted). Thus, providing a general symbolic framework for disassembly sequence planning under uncertainty. The main contributions of this work are:\\
We introduce a general POMDP formulation of disassembly planning that explicitly models end-of-life conditions that affect the disassembly process as hidden variables.
We develop a system that automatically generates specific POMDP models from CAD data, robot capabilities, and prior inspection data, and solves them to obtain instance-specific disassembly policies. We show in simulation that these POMDP-based policies outperform deterministic baselines in terms of both average disassembly time and variance across three different products shown in Fig.~\ref{fig:bg3-motor}. We also show that the system generalizes to two different robotic setups with varying capabilities showing successful disassembly of two different products.

\begin{figure*}[ht]
\centering
\includegraphics[width=\textwidth]{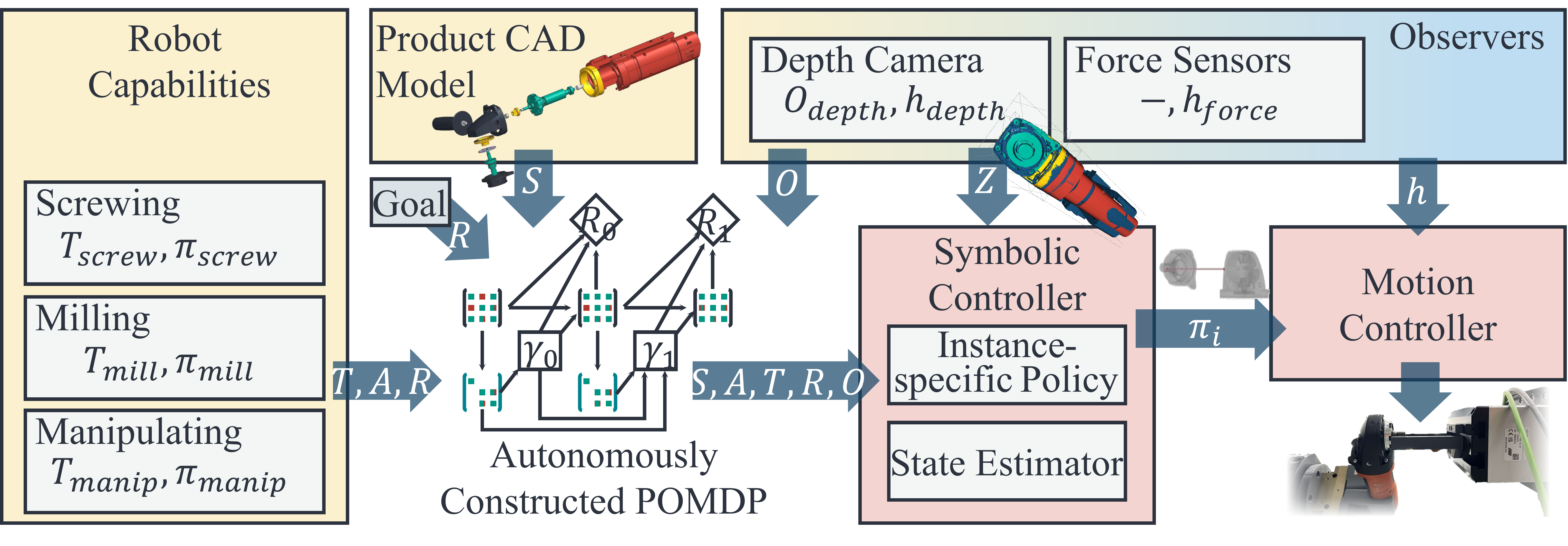}
\caption{Block diagram of the proposed system colored according to sensing (blue), planning (yellow), and execution (red) components. Disassembly is decomposed into a symbolic task planner based on a POMDP formulation and a motion planner that generates executable trajectories.}
\label{fig:system-overview}
\end{figure*}
\section{RELATED WORK}\label{sec:related-work}
While uncertainty has been considered in previous disassembly sequence planning research, it was mostly treated at an aggregated, factory-control level across many products. As a result, partial observability at the level of individual products and operations has not yet been explored~\cite{dissasembly-review}.
However, disassembly sequence planning  is essentially a domain-specific form of task planning, where the objective is to determine the sequence of actions that removes some or all parts from a given assembly.
It is thus natural that our approach draws on the task-and-motion planning (TAMP) literature such as~\cite{tampura,tamp-failing}.
However, the focus of this work is not the planning algorithm (indeed, we will later see that even a certainty equivalent controller already improves on the current state of the art), but the formulation of disassembly as a POMDP and its automatic generation from CAD data. A range of symbolic models have been proposed to represent the structure of a product, including graph-based approaches \cite{kartik},  Disassembly Petri Nets \cite{zhao2014fuzzy}, and matrix-based precedence models \cite{smith2011rule}. However, they typically encode binary part-connection states (connected/disconnected) and therefore cannot express richer EOL conditions such as wear or corrosion. A recent work  \cite{malte} proposed a more expressive connectivity relation that can capture the influence of destructive action, such as milling, on disassembly. While this work did not consider end-of-life conditions or uncertainty, we argue that end-of-life conditions can be seen as a form of destruction and thus use the same state representation while extending the model to handle uncertainty.

\section{System Architecture}\label{sec:system-architecture}
The proposed robotic disassembly system combines a symbolic task planner, utilizing a POMDP as its system model, with a motion planner that executes the resulting actions (Fig.~\ref{fig:system-overview}). The POMDP is constructed from what we call capabilities and observers, which are abstractions of the tools and sensors available in the robotic system.
A capability is defined as the ability of a given system to perform a certain task or process, following \cite{control-architecture}.
In this work, we define a capability as a tuple $(T_k,\pi_k)$ where $\pi_k$ is a motion policy or trajectory planner associated with the tool $k$ and $T_k$ is the transition function that describes how the symbolic state changes when using the tool $k$. The necessary information to estimate the symbolic state and parameterize the motion policies and trajectory planners is provided by the observers.
These are defined as a tuple $(O,h)$ where $O$ describes the observation function of the symbolic state and $h$ is the measurement function that provides the motion planners with the necessary information to generate the motion. These include, but are not limited to, the pose of the component to be removed, the pose of the robot, or force feedback.
The execution of both the symbolic and motion planning is supervised by the symbolic and motion controllers.
Since the POMDP is the central component of the system, the next section will present a general formulation of disassembly as a POMDP.
This formulation is then grounded in product, capability, and observer models in Sec.~\ref{sec:system-implementation}.

\section{A POMDP formulation of disassembly}\label{sec:pomdp}
A POMDP is defined by a tuple 
$\langle S, A, T, R, O, Z, \gamma \rangle$, 
where $S$ is the set of states, $A$ the set of actions, 
$T$ the transition function, $R$ the reward function, 
$O$ the observation model, $Z$ the set of observations, 
and $\gamma$ the discount factor.
The next subsections will now describe how each of these components can be defined to model disassembly planning under uncertainty.

\subsubsection{The state space $S$} \label{sec:state-space}
\begin{figure}[t]
\centering
\includegraphics[width=1\columnwidth]{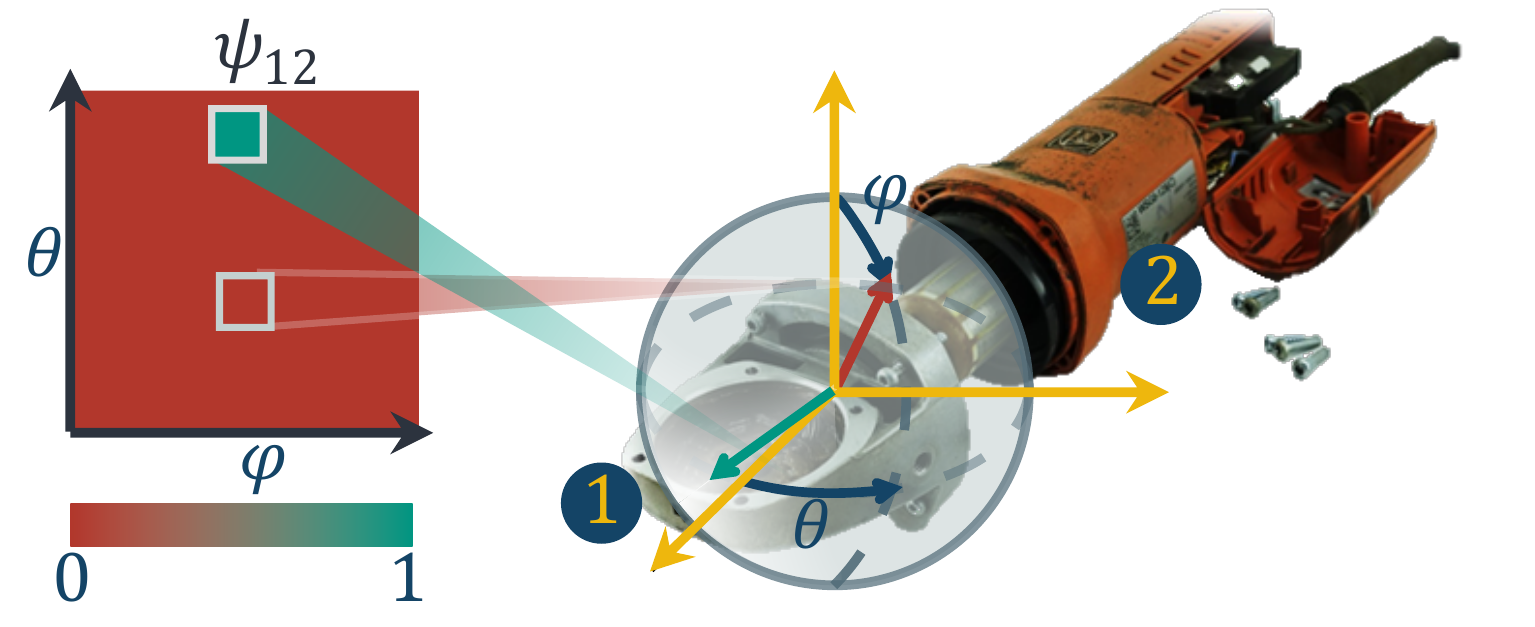}
\caption{Surface plot of the disassembly relation $\psi_{12}(\theta,\phi)$ between the head (1) and the body (2) of an angle grinder. $(\theta,\phi)$ describe the movement direction in spherical coordinates $(\theta,\phi)$. The  head can only be pulled in one direction (modelled as $\psi_{12}(\theta,\phi)=1$) resulting in a single green area.}
\label{fig:psi-visualization}
\end{figure}

For disassembly planning, the state should represent the product structure and how its parts can be separated. In \cite{malte}, this is described using directional relations between two parts $i$ and $j$ denoted $\psi_{ij}(\theta,\phi)$. If part $i$ can be moved relative to part $j$ in a spatial direction $(\theta,\phi)$ then $\psi_{ij}(\theta,\phi)=1$, otherwise $\psi_{ij}(\theta,\phi)=0$ (shown in Fig.~\ref{fig:psi-visualization}). While this directional encoding captures the ground truth of whether two parts can be moved relative to each other, it is inherently deterministic and does not account for uncertainty. To capture this, we reinterpret $\psi_{ij}(\theta,\phi)$ as the probability that $i$ can be moved relative to $j$ in direction $(\theta,\phi)$. When uncertain whether two parts are stuck together, we may set $\psi_{ij}(\theta,\phi)=0.5$ where a nominal model would use $1$.
The binary ground truth of $\psi_{ij}(\theta,\phi)$ is derived from CAD using the algorithm of \cite{malte-product-model}, which analyzes contact points between parts. Since this approach is sensitive to small gaps or corner contacts, we adjust this algorithm by incorporating ray-casting to more robustly detect contacts.
EOL conditions can be viewed as the result of actions that were implicitly applied to the product state during its lifetime. Formally, each condition corresponds to an operator $f_c$ that modifies the nominal disassembly relation $\psi^0_{ij}$ obtained from CAD, yielding the actual relation $\psi_{ij}=f_c(\psi^0_{ij})$. In this sense, corrosion, wear, or deformation can be treated analogously to destructive disassembly actions in \cite{malte}, which occurred over the lifetime of the product. This interpretation allows us to integrate EOL variability seamlessly into the same symbolic framework: the initial state of the POMDP is simply the CAD-derived structure transformed by a (hidden) sequence of such EOL operators.
Table \ref{tab:part-connection-effects} shows how different EOL conditions could be represented in this model.
The full product state $\Psi$ is now a matrix comprising all pairwise disassembly relations $\psi_{ij}(\theta,\phi)$ between parts $i$ and $j$.
While in \cite{malte}, removed parts were deleted from the state we opted for a fixed state dimension to simplify the POMDP formulation.
Removed parts are thus represented by $\psi_{ij}(\theta,\phi)=1~\forall j$.
\begin{table}[h]
\centering
\begin{tabularx}{\columnwidth}{XXX}
\toprule 
\textbf{EOL Condition} & \textbf{Consequence for Disassembly} & \textbf{Operator $f_c(\psi^0)=...$} \\
\midrule
Fretting Corrosion, Aged Adhesive & Blocked in all directions &$ 0 $ \\
Wear or Loosening & Possible in additional directions & $ \psi^0 +\psi_{new}(\theta,\phi)$ \\
Plastic Deformation & Directions changed & $ \psi^0(\theta+\Delta\theta, \phi+\Delta\phi)$ \\
Structural Damage & Arbitrary & $\psi_{new}$ \\
\bottomrule
\end{tabularx}
\caption{Example EOL conditions and how they could be modeled using $\psi_{ij}$.}
\label{tab:part-connection-effects}
\end{table}
While $\Psi$ encodes geometric relations, tool requirements are modeled separately via the tool matrices $C_k$ adapted from \cite{malte}, which specify which part pairs can be separated with tool $k$. 
These matrices serve as filters on the action space rather than additional state variables.
Similarly, while the full state $\Psi$ is required to model general end-of-life conditions, a lower-dimensional abstraction can be used for decision making, depending on the specific EOL conditions considered.
One such example will be presented in section \ref{sec:symbolic-planning}.
\begin{figure*}[h]
        \centering
        \includegraphics[width=\textwidth]{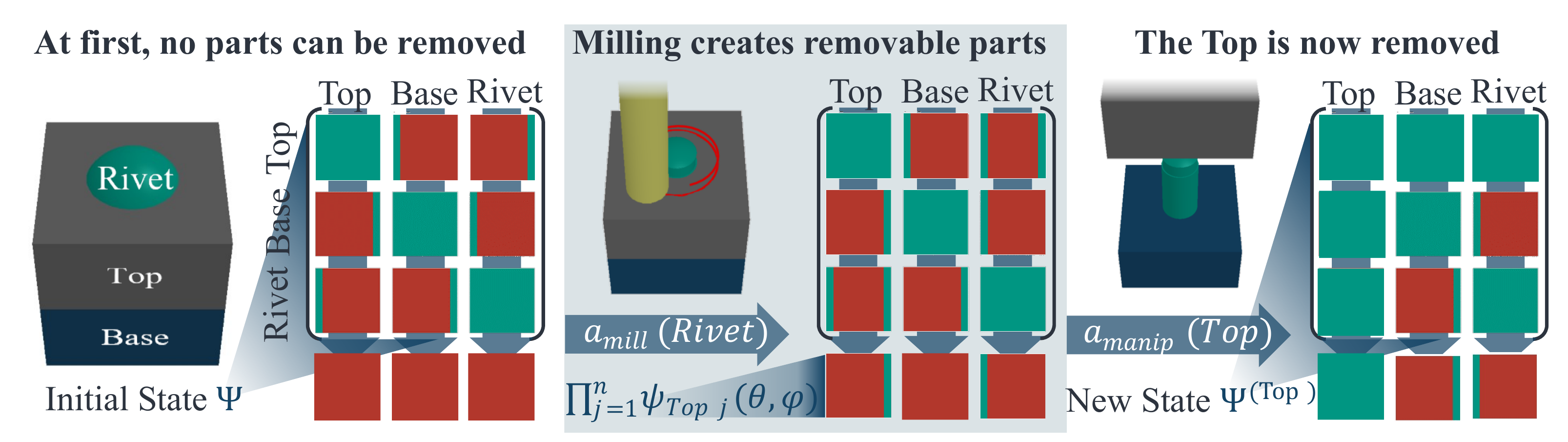}
        \caption{Illustration of how transition function $T$ changes $\Psi$ (individual $\psi_{ij}(\theta,\phi)$ shown as surface plots) using a rivet as example. 
 Initially, $\prod_{j=1}^{n} \psi_{ij}(\theta,\phi)$ is zero (red in each surface plot) for all directions $(\theta,\phi)$, meaning neither the rivet nor the top can be removed non-destructively. 
 After applying the milling action $a_{mill}$, the relations $\Psi_{Rivet,Top}$ and $\Psi_{Top,Rivet}$ now feature new directions where $\prod_{j=1}^{n} \psi_{ij}(\theta,\phi)=1$ (green),
 enabling removal of either part with a manipulation action $a_{manip}$. Disassembled parts are represented by $\psi_{ij}=1~\forall j$.}
        \label{fig:transition-function-illustration}
\end{figure*}

\subsubsection{The action space $A$}
For end-of-life disassembly, relying only on standard manipulating or screwing is insufficient: a rusted screw may prevent disassembly entirely if no alternative paths to reach the target parts are available. Prior work \cite{disassembly-sequence-planning-partial-destructive-mode,backup} addressed this by adding destructive actions (cutting, drilling, breaking), which ensure progress when non-destructive actions fail. In \cite{malte}, this principle was generalized by modeling destructive actions as operators on the product structure itself. We build on this view and integrate destructive actions into our POMDP formulation.
A destructive action on part $i$ modifies the disassembly relation $\psi_{ij}(\theta,\phi)$ for all $j$, effectively creating new feasible directions that allow $i$ to be removed. While such actions could in principle be continuously parameterized, we follow the approach of \cite{malte} and constrain them to discrete operators that directly enable the removal of $i$. This yields at most one destructive action per part, simplifying the space while remaining more general than \cite{disassembly-sequence-planning-partial-destructive-mode,backup}, since it also covers cases such as cutting holes to free a part that would otherwise require multiple steps to remove.
Non-destructive actions are defined separately for each tool $k$, yielding discrete sets $A_k$ of actions $a_k(i)$ (removing part $i$ with tool $k$). The overall action space is $A=\bigcup_k A_k$. Assuming each part can only be removed with one non-destructive tool, the size of $A$ equals the number of parts.
For practical reasons, we may filter out actions with near-zero likelihood of changing the state.
This can be expressed using the following prerequisite function $preq(a_k(i))$:
\begin{equation}
preq(a_k(i))  := T(\Psi^{(i)}| \Psi,a_k(i)) > \delta
\end{equation}
where $T$ is the transition function describing the probability of removing part $i$ given the current state $\Psi$ and action $a_k(i)$. $\delta$ is a certainty threshold that describes how certain the agent must be that the part $i$ can be removed. For this work, $\delta=0$ was used due to the limited number of parts.
This prerequisite function only models the geometric and symbolic constraints of the product. Meaning it describes whether a part can be moved in principle.
This does not mean that any particular robotic system is actually able to perform the action at the motion level. For example, if the part is located in a very confined space, the robot might not be able to reach it.
For this reason, all actions undergo a secondary feasibility check that uses the motion policy $\pi_k$ of the tool $k$ to determine whether the action can actually be performed.
Since this is very expensive, we adopt the iterative feasibility check approach presented in \cite{iterative-feasibility}, which first checks whether the tool can reach the part, then checks whether the robot can reach the part, and so on, before actually evaluating the full motion policy. 

\subsubsection{The transition function $T$}
The transition function $T$ follows directly from the action space and describes the probability of moving from one state to another given an action.
For a non-destructive action $a_k(i)$, we define the transition function and its two possible outcomes as:
\begin{equation}
\begin{aligned}
T(\Psi^{(i)}| \Psi,a_k(i)) &= p_{success}\max_{\theta,\phi}\prod_{j=1}^{n} \psi_{ij}(\theta,\phi) > 0\\
T(\Psi| \Psi,a_k(i)) &= 1 - T(\Psi^{(i)}| \Psi,a_k(i))\\
\end{aligned}\label{eq:transition-function}
\end{equation}
Where $\Psi^{(i)}$ is the $\Psi$ part $i$ removed as shown in section \ref{sec:state-space} and $p_{success}$ is the probability that the motion policy $\pi_k$ of tool $k$ will successfully execute the action $a_k(i)$.
Since destructive actions such as milling always remove material regardless of whether it is present, we model them with a deterministic transition function:
\begin{equation} \label{eq:transition-function-destructive}
T(\psi^{new}_{ij}(\theta,\phi)| \Psi,a_{mill}(i)) = 1
\end{equation}
where $\psi^{new}_{ij}(\theta,\phi)$ is the new disassembly relation after the destructive action $a_{mill}(i)$ has been applied.
This notation assumes that the destructive action does not create any new parts. However, if one wants to cut a hole into a part, 
it might make sense to simply cut out a piece that can be removed non-destructively. 
This can be modeled by expanding $\Psi$ to include the newly created parts. 
Since this would change the dimensionality of the state space,
and one would often immediately want to remove the newly created part anyway,
we instead opted for defining both cutting out a piece and removing it as a single destructive action.
An illustration of the transition function operating on the ground truth product state is shown in Fig.~\ref{fig:transition-function-illustration}.
\subsubsection{The observation function $O$}
There is no general formulation for the observation function $O$ like there is for the transition functions $T$. The reason is that the observation function strongly depends on the type of sensors, as well as the algorithm used to map sensor data onto the symbolic state.
This section will therefore only provide some classification of different observer types.
A detailed overview of the observation functions implemented in the robot cells can be found in section \ref{sec:capabilities}.
In general, we differentiate between two types of observation functions.
Action feedback observations, $O_a(\hat{\Psi}| \Psi,a_k(i))$ are performed after execution of an action and direct state observations, $O_s(\hat{\Psi}| \Psi)$ which do not depend on any action.
A example of action feedback observations are checking whether a part that should be movable in a certain direction could actually be removed.
This gives us some feedback about how the internal structure might deviate from the original CAD model. For example, sometimes two parts might be glued together internally.
Even if we don't see the second part, pulling at the first could help us determine this information.
In this case, the observation could be expressed as
\begin{equation}
\begin{aligned}
O_{glue}(\hat{\Psi}^{(i)}| \Psi,a_k(i))  & = \max_{\theta,\phi}\prod_{j=1}^{n} \psi_{ij}(\theta,\phi)\\
O_{glue}(\hat{\Psi}| \Psi,a_k(i)) & = 1-O_a(\hat{\Psi}^{(i)}| \Psi,a_k(i))\\
\end{aligned}
\end{equation}
Where $\hat{\Psi}^{(i)}$ is the estimated state after part $i$ has been removed and $\hat{\Psi}$ is the estimated state if part $i$ could not be removed.
This requires some type of sensor to check whether the part has been removed.
This could, for example, be done optically, or using a force sensor that detects that the gripper simply slipped instead of pulling out the object.
By contrast, direct state observations $O_s(\hat{\Psi}^{(i)}| \Psi)$ provide feedback about the state irrespective of any action taken.
For example, if a camera is used to observe the product, a classifier model could be used to detect whether a screw is rusted or not.
The observation function can then be defined as:
\begin{equation}
 O_{rust}(\hat{\psi}_{ij}=0 | \Psi) = \begin{cases}
 p_{TP} & \text{if } \exists (\theta,\phi): \psi_{ij}(\theta,\phi) = 0\\
 p_{FN} & \text{if } !\exists (\theta,\phi): \psi_{ij}(\theta,\phi) = 0\\
        \end{cases}
\end{equation} 
Where $p_{TP}$ is the true positive rate and $p_{FN}$ is the false negative rate of the classifier as determined by the training data.

\subsubsection{The reward function $R$}
The goal of the presented system is not necessarily the complete disassembly of the product, but rather the removal of a predetermined set of valuable parts in the shortest time possible.
The time for each action can be measured using the motion policy $\pi_k$ of the robot computed during the action feasibility check.
It includes the time to actually remove the part, but also the time incurred to change the end-effector, and the time to place the removed part in a bin and move to the next part. Since we don't want the system to just destroy every part, a part-specific penalty term $r_{value,i}$ is added to the reward function that describes the value of each component. This is meant to discourage the system from blindly cutting its way to the target parts.
The full reward function is then defined as:
\begin{equation}
R(\Psi,a_k(i)) = -\left( t_{action} + t_{change} + t_{place} + r_{value,i} \right)
\end{equation}
where $t_{action}$ is the time to perform the action, $t_{change}$ is the time to change the end-effector (if required), and $t_{place}$ is the time to place the removed part in a bin.

\section{System Implementation}\label{sec:system-implementation}
\begin{figure*}[ht]
\centering
\includegraphics[width=\textwidth]{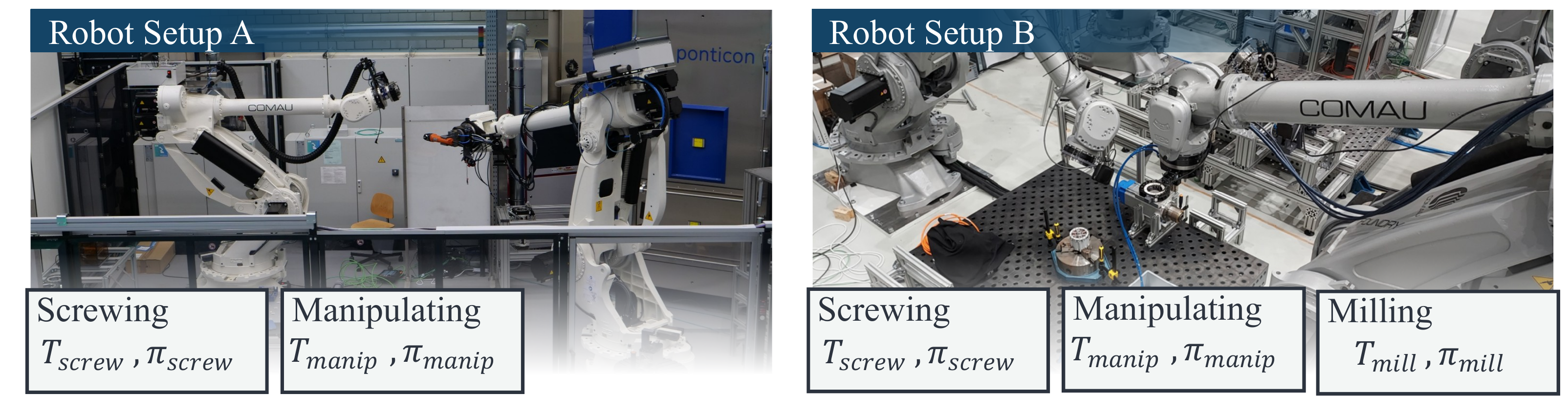}
\caption{The two robotic systems used for the experiments with their respective capabilities.}
\label{fig:robotic-systems}
\end{figure*}

We now describe the implementation of the architecture introduced in Sec.~\ref{sec:system-architecture}. 
This involves grounding the symbolic POMDP formulation in specific product, capability, and observer models as well as the reinforcement learning algorithm used to solve the POMDP.
Specifically, the POMDP state and transitions are instantiated from CAD models and initial inspection data; the observation and measurement functions ($O,h$) are realized through vision- and force-based observers; and the transition and policy pairs ($T,\pi$) are defined for each available robotic capability. 
Together, these components implement the full pipeline from CAD to executable disassembly actions.
Two systems are used in this work with different capabilities, but sharing the same observer (Fig. \ref{fig:robotic-systems}).
System~A supports manipulation and screwing; System~B also supports milling.
\subsection{Extracting the Product Model from CAD}
The base representation of the product structure is extracted from the CAD model using the method described in \ref{sec:state-space}.
In this work, the end-of-life condition considered is that parts are stuck, for example, due to corrosion or adhesive.
This is modeled according to the first row of Table~\ref{tab:part-connection-effects} as a disassembly relation $\psi_{ij}(\theta,\phi) = 0 \forall j$ for all parts $i$.
We assume that there is an initial inspection step as described in \cite{control-architecture,dominik} that provides an initial belief over the product state.
In practice, this belief could also be inferred from prior disassembly experiences or a combination of both.
\begin{figure}[h]
\centering
\includegraphics[width=\columnwidth]{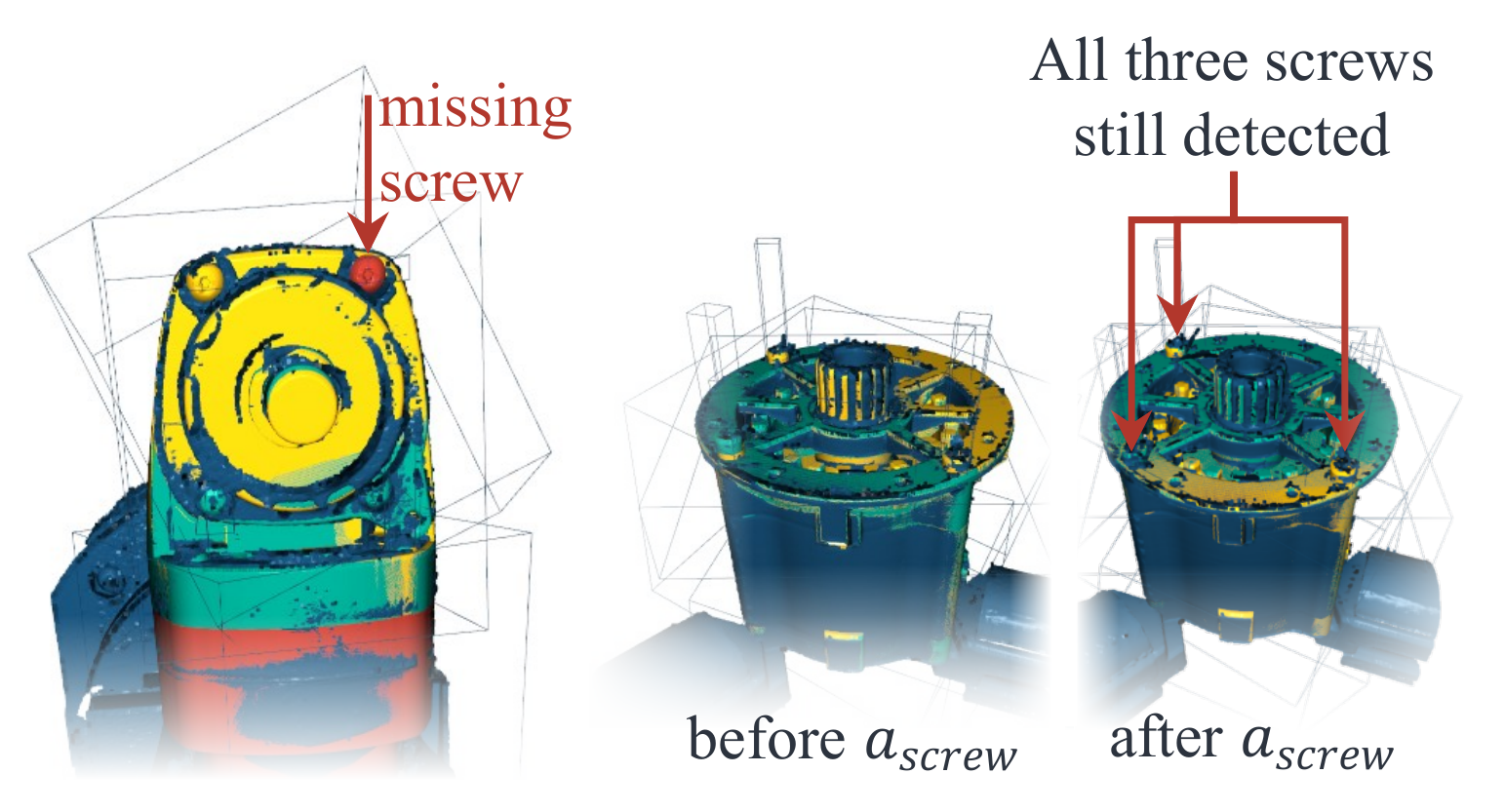}
\caption{Sample results of the state estimation process for both the angle grinder and electric motor from \cite{malte}. 
The real point cloud (blue) is compared with the simulated point cloud (yellow).
The latter is then repositioned to align with the real point cloud (shown in green). This is used to determine $O_{missing}$(red).
Parts that are present before and after an attempted removal are used to determine $O_{stuck}$ (shown on the right).}
\label{fig:estimation-visualization}
\end{figure}
\subsection{Observers} \label{sec:observers}
Both systems share a Zivid depth camera combined with the point-cloud matching method from \cite{malte-zivid}.  
CAD models are inserted into a simulation of the robot cell to obtain synthetic depth images, which are aligned with real point clouds to localize assemblies and estimate part poses (Fig.~\ref{fig:estimation-visualization}).  
This process defines the measurement function $h$, which provides the motion planners with the required continuous information about the  6D part poses.
The observation function $O$ used in the symbolic POMDP is derived from these measurements. In other words, $O$ is not measured directly, but obtained by mapping the result of $h$ into symbolic statements about the disassembly state. For example, if $h$ estimates that the pose can not be found because it likely does not exist, the corresponding symbolic observation is ``part missing,'' expressed as $\psi_{ij}(\theta,\phi)=1~\forall j$. If $h$ detects that the pose of a part has not changed after an attempted removal, $O$ encodes this as ``part stuck,'' i.e.\ $\psi_{ij}(\theta,\phi)=0~\forall j$.  
Formally, the two main observation models derived from $h$ are:
\begin{equation}
O_{stuck}\big(\hat{\psi}_{ij}=0~\forall j \mid\Psi, a_k(i)\big) =
\begin{cases}
p_{\mathrm{TP}} & \text{if } ||h_{new}-h_{old}|| < \epsilon,\\
p_{\mathrm{FP}} & \text{else}.
\end{cases}
\end{equation}
\begin{equation}
O_{missing}\big(\hat{\psi}_{ij}=1~\forall j \mid \Psi\big) =
\begin{cases}
p_{\mathrm{FN}} & \text{if } h \text{ detects part},\\
p_{\mathrm{TN}} & \text{ else}.
\end{cases}
\end{equation}
Following \cite{malte-zivid}, we assume $p_{\mathrm{TP}}=1$, $p_{\mathrm{FP}}=0$, $p_{\mathrm{FN}}=0$, and $p_{\mathrm{TN}}=1$. This idealization ignores lighting effects that may induce false negatives, which we leave for future work. 
A second observer is implemented that provides force feedback $h_{force}$ for motion policies that use force control.
In principle, it could also be used to detect rusted screws using the approach presented in \cite{mangold2023systematic}. However, since the depth camera already provides more general information about arbitrary stuck parts, we did not implement a symbolic observation function for the force observer.

\subsection{Capabilities}  \label{sec:capabilities}
Each end-effector $k$ of the robotic system defines a capability $(T_k,\pi_k)$. 
The robots are capable of autonomous tool changes, allowing them to switch between different end-effectors during disassembly.
To specify the capabilities, the transition function $T_k$ and the motion policy $\pi_k$ have to be defined for each tool,
as well as the reward incurred by each action.
For non-destructive actions, $T_k$ follows (\ref{eq:transition-function}).
The motion policy for the gripper is a straight pulling motion along the direction $(\theta,\phi)$ that maximizes $\psi_{ij}(\theta,\phi)$, 
with a fixed pulling distance of 100 mm, followed by a transport to a drop-off position.
For the screwdriver, the motion policy is a force-controlled spiral screw search with subsequent torque-controlled unscrewing as implemented based on~\cite{screw-search}.
The success probability $p_{success}$ was determined experimentally, as well as the action time $t_{screwing}$.
All other time components $t_{change}$ and $t_{place}$ were measured using a simulation of the robot cell.
The milling tool defines a destructive capability with $T_{\text{mill}}$ as in (\ref{eq:transition-function-destructive}). 
Its policy $\pi_{\text{mill}}$ follows \cite{malte}: the target part is projected onto the obstructing geometry to derive the material to be removed, and a CAM-generated toolpath is executed. 
Execution time depends on toolpath length and spindle feed.
Since motion-level computations are costly, we precompute a POMDP graph for each product, parameterized by the initial belief over EOL conditions. 
This graph can be reused across multiple training runs, substantially reducing computation.

\section{Symbolic Controller} \label{sec:symbolic-planning}
The symbolic controller selects the next action using the POMDP dynamics introduced in Sec.~\ref{sec:pomdp}. To obtain a tractable solution, we approximate the belief space by marginalizing the latent EOL conditions with their inspection priors, resulting in an expectation-MDP whose transition and reward models are derived from the POMDP. A Q-learning agent ($\gamma=0.99$, initial learning rate $0.7$ with decay, $5000$ episodes per product) is trained on this model to minimize expected disassembly time. 
At execution time, a Bayesian filter updates beliefs as new observations arrive using the observation functions defined in Sec.~\ref{sec:observers},
The controller then acts on the most likely state. This design sacrifices explicit knowledge-seeking behavior: the agent will not choose exploratory actions solely to reduce uncertainty. However, since action feedback immediately reveals the hidden state, and backup options exist to recover from failures, this was deemed acceptable since it already addresses the main weaknesses of deterministic planners.
It reasons under uncertainty over the end-of-life conditions rather than assuming brand new parts, adapts actions as new evidence becomes available, and maintains robustness against unexpected failures. However, the agent represents only a basic first instantiation of this framework; more sophisticated belief-dependent controllers can be substituted directly and are expected to yield further gains.

\section{Motion Controller} \label{sec:motion-controller}
The motion controller is structured around the digital twin, which integrates descriptive, prescriptive, and predictive models  \cite{wortmann} into a single control pipeline. The descriptive model represents the current poses of all parts estimated via the observer function $h$, as well as which tool is currently being used. 
The prescriptive model specifies the tool policies $\pi_k$ that define how each tool $k$ operates for a given disassembly action. 
The predictive model simulates the application of these policies using a simulation of a robot and assembly, rolling out behaviors such as a pulling motion of the part along $(\theta,\phi)$ determined by $ \max_{\theta,\phi} \psi_{ij}(\theta,\phi)$
or collision-free paths between assembly and drop-off or tool change positions.
The result of this rollout is G-code, which not only encodes deterministic robot trajectories, but also includes adaptive behavior such as the screw search described in section \ref{sec:capabilities},
or force feedback monitoring. The loop of the digital twin is closed during the next observation cycle, where the observer function $h$ updates the descriptive model with new sensor data.

\section{Experiments}\label{sec:comparison}
\begin{figure}
\centering
\includegraphics[width=\columnwidth]{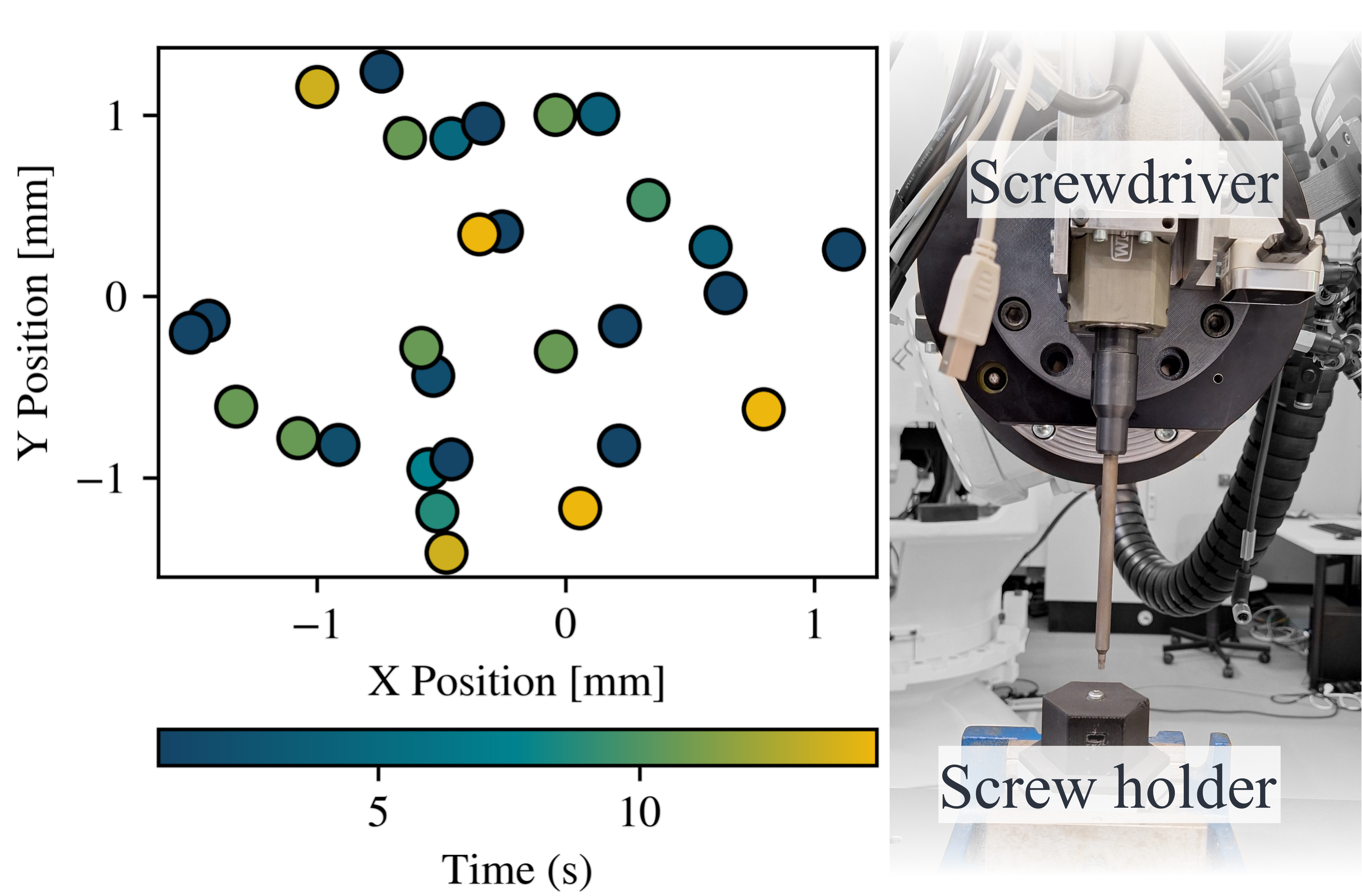}
\caption{Measured time to execute the screw action as a function of the initial offset between screwdriver and screw head. Experiment setup shown on the right.}
\label{fig:action-times}
\end{figure}

\begin{figure}[h]
\centering
\includegraphics[width=\columnwidth]{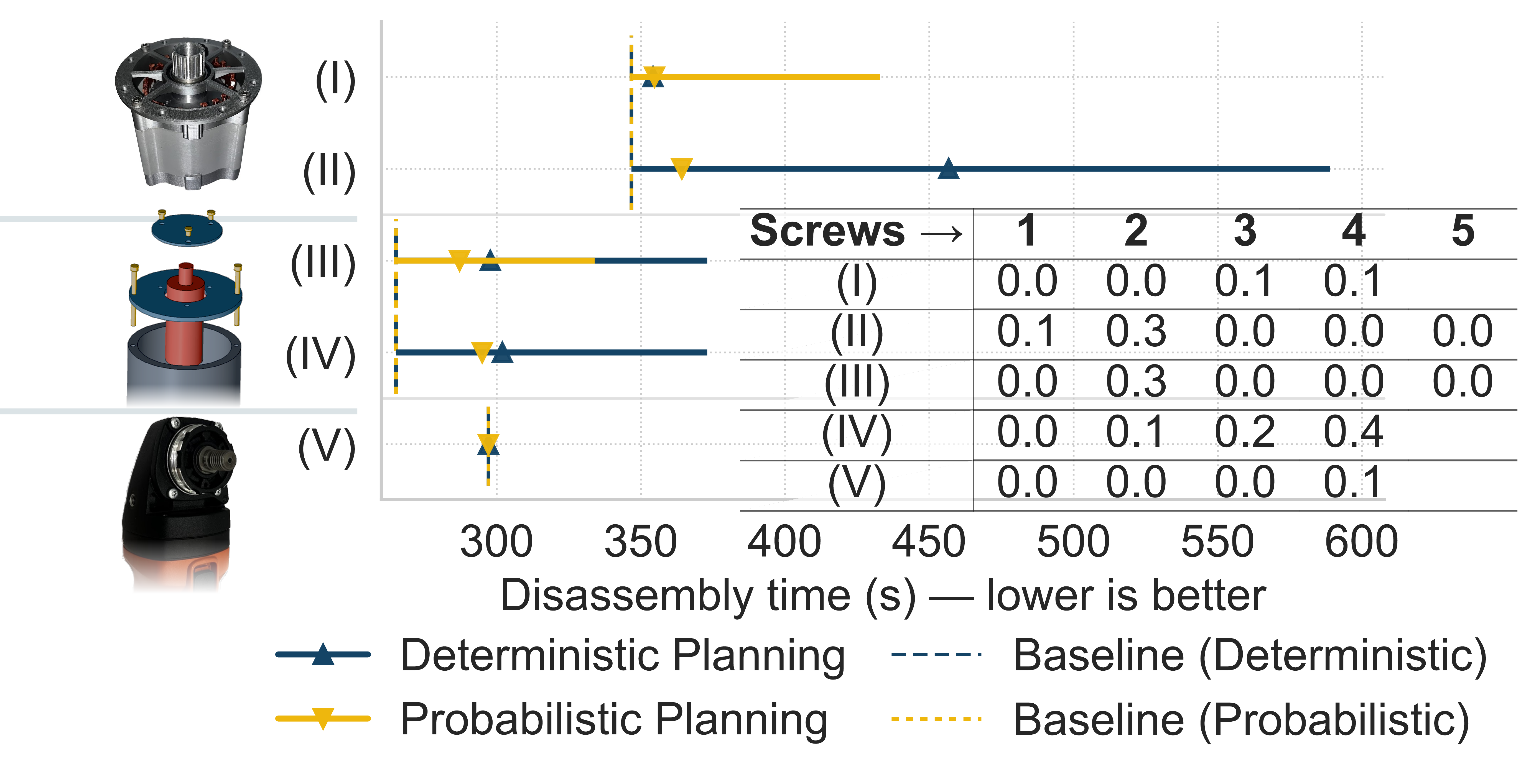}
\caption{Disassembly times for different products with increasing probabilities that screws are stuck (shown in the table). For the second product, screws 1–2 belong to the large lid and screws 3–5 to the small lid.
Probabilistic planning pulls ahead of deterministic planning as uncertainty increases (I-IV). If no alternative strategies are available (V), both planners perform equally well.}
\label{fig:reward-functions}
\end{figure}

\begin{figure*}[h]
\centering
\includegraphics[width=\textwidth]{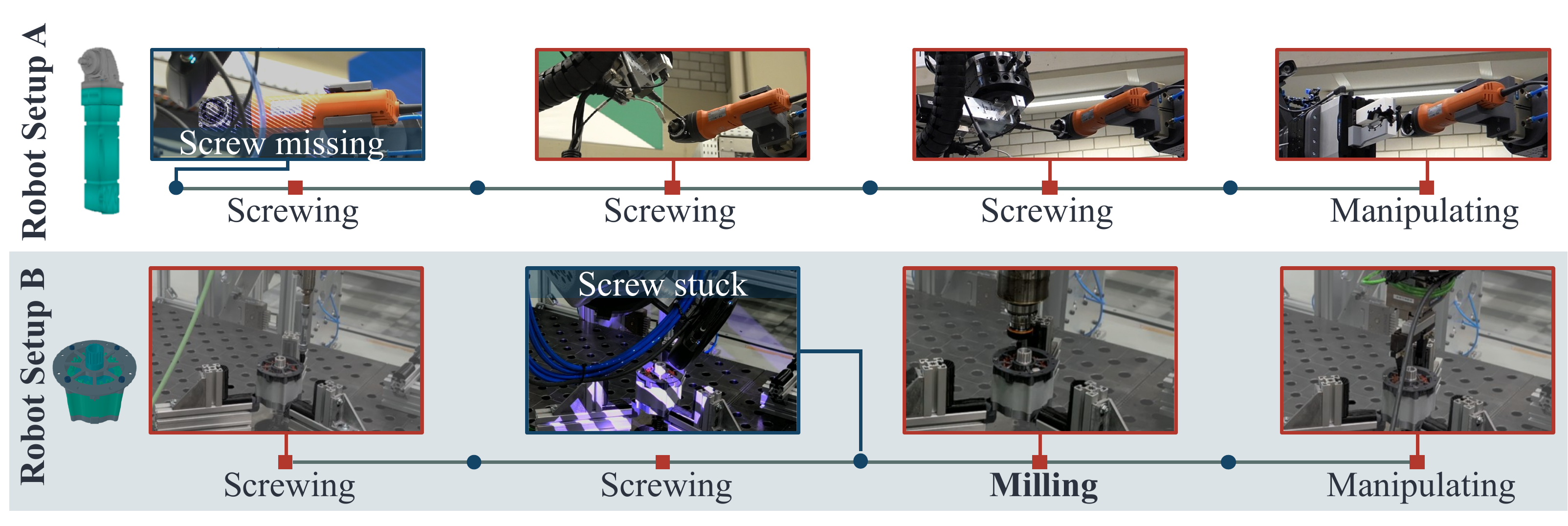}
\caption{Rollout of the probabilistic planner on real hardware for the electric motor and angle grinder. The sequence shows the executed actions (red squares) and corresponding observations (blue circles). The agent successfully identifies and handles the EOL conditions (stuck screw and missing screw) using the observations provided by the observer function $O$ (Fig.~\ref{fig:estimation-visualization}).}
\label{fig:execution-results}
\end{figure*}

The first experiments were designed to parameterize success probabilities for the non-destructive actions as well as the time for the screw search.
Since these policies are deterministic, we assumed that the only source of uncertainty is the finite accuracy of the measurement function $h$ 
used to localize the parts. Based on a position variance reported in \cite{malte-zivid}, two experiments were set up.
For the manipulation action, this involved trying to remove each part 30 times with a normally distributed offset according to the pose variance.
Since the policy managed to remove every part in all 30 attempts, we set $p_{success,manipulation} = 1$.
A similar setup was designed for the screwing action, where a screw had to be unscrewed 30 times with the same normally distributed offset.
Without the screw search, the system achieved a very low success rate of $p_{success,screwing} = 0.6$. 
However, with the screw search implemented, the system managed to unscrew every screw in all 30 attempts, leading to $p_{success,screwing} = 1$.
The measured times to execute the screw action as a function of the initial offset between screwdriver and screw head are shown in Fig.~\ref{fig:action-times}.
For the POMDP, we conservatively use the maximum measured duration.
Using parameterized transition functions and action times, we now evaluate the performance of the probabilistic planner against a deterministic planner on different products.
Specifically, we want to investigate our claim that probabilistic planning leads to more robust disassembly plans under EOL uncertainty.
To this end, we evaluated the performance of our probabilistic planner on three products: the electric motor from \cite{malte}, the electric motor with two lids, and an angle grinder, all shown in Fig.~\ref{fig:bg3-motor}. The disassembly goal was to remove the rotor for the motors and the gearbox for the angle grinder.
The electric motor was tested on the capabilities of robotic system B, while the electric motor with two lids and the angle grinder were tested on robotic system A (both shown in Fig.~\ref{fig:robotic-systems}).
Each product was now tested under different EOL conditions. Brand new parts, with no conditions, as well as parts where one or more screws were rusted and thus stuck with increasing probability.
The resulting reward distributions are shown in Fig.~\ref{fig:reward-functions}.
In cases without EOL conditions, both planners perform equally well. Since the deterministic planner is mathematically optimal in this case, the probabilistic planner can at best match its performance. As uncertainty is introduced,
the probabilistic planner starts to pull ahead. For the electric motor from \cite{malte}, the system at some point starts to simply mill the lid off directly, resulting not just in a higher expected reward, but also a zero variance since this action always works. Something similar can be observed for the electric motor with two lids. Initially, the probabilistic planner at least tries to unscrew the screw that might be rusted. This is not explicit knowledge-seeking behavior,
which would require a belief-dependent planner, but rather a side effect of the Q-learning approach. If the screw is not rusted, unscrewing it is faster than milling the lid off. However, if it is rusted, milling the lid off is faster than trying to unscrew the rusted screw and then milling the lid off anyway.
In scenario (IV) it then commits to disassembling the smaller lid instead. For the angle grinder, no difference in the strategy can be observed. The reason for this is simple: since robot cell A does not have a milling tool, there are no alternative strategies to disassemble the gearbox if the screw is rusted. On the one hand, one could thus conclude that in such situations, a probabilistic planner is not needed. However, the benefit of a probabilistic description is that the system can still report the probability of success on a given plan.
If one thus has both robotic systems available, one could use the POMDP model to determine which products should be disassembled on which system in order to maximize the overall throughput of the disassembly facility. In summary, we can conclude that with alternative strategies available, the probabilistic planner achieves faster disassembly times with lower variance under EOL uncertainty, and in the worst case, matches the performance of a deterministic planner.
To test the whole pipeline from capabilities and CAD data to physical disassembly, both the electric motor from \cite{malte} and the angle grinder were disassembled on the physical robotic systems.
To simulate different EOL conditions, one screw of the electric motor was stuck (simulated by the experimenter screwing it back in after the fact), and one screw of the angle grinder was initially missing.
The resulting behavior is shown in Fig.~\ref{fig:execution-results}. 
In the case of the electric motor, the agent first decides to unscrew every screw; however, after finding one screw to be stuck, as shown in Fig.~\ref{fig:estimation-visualization},
it decides to instead mill the lid off and then remove the rotor. The agent also correctly identified that one screw is missing (Fig.~\ref{fig:estimation-visualization}) and thus did not waste time trying to unscrew it.
The motion planner was able to successfully execute all actions using the parameterized motion policies.
We can thus conclude that the pipeline is not only able to plan more robustly under EOL uncertainty, but also able to execute the plan on a physical robotic system.

\section{CONCLUSION} \label{sec:discussion}
We presented a robotic system that formulates disassembly planning as a POMDP derived automatically from CAD data, robot capabilities, and inspection results. 
By explicitly modeling hidden EOL conditions such as stuck or missing parts, our approach yields more robust task plans than deterministic baselines, as shown across multiple products and robotic setups. 
Our current controller uses a reinforcement-learning approximation with stochastic action outcomes informed by inspection priors, while a Bayesian filter tracks beliefs during execution. 
Although this certainty-equivalent strategy does not yet exploit the full belief space, it already improves robustness and reduces variance. 
Future work will investigate belief-dependent solvers and richer observation models, enabling more adaptive and information-seeking disassembly strategies.

\bibliographystyle{IEEEtran}
\bibliography{IEEEabrv,bibliography}

\end{document}